\documentclass[a4paper,fleqn]{cas-dc}

\usepackage[numbers]{natbib}
\usepackage{amsmath}
\usepackage{graphicx}
\usepackage{amsfonts}
\usepackage{amssymb}
\usepackage{booktabs}
\usepackage{colortbl}
\usepackage{paralist}
\usepackage{booktabs}
\usepackage{multirow}
\usepackage{soul}
\usepackage{amsmath}    
\usepackage{multicol} 
\usepackage{float}
\usepackage{caption}
\usepackage{subcaption}
\usepackage{perpage} 
\usepackage{soul} 
\usepackage{xcolor}   
\usepackage{ulem}     
\usepackage{commath}
\usepackage{tablefootnote} 
\usepackage{pifont}
\newcommand{\cmark}{\ding{51}}%
\newcommand{\xmark}{\ding{55}}%
\usepackage{textcomp}
\usepackage{hyperref}
\hypersetup{hidelinks=true}
\def\tsc#1{\csdef{#1}{\textsc{\lowercase{#1}}\xspace}}
\tsc{WGM}
\tsc{QE}

\MakePerPage{footnote} 
\begin{document}
\let\WriteBookmarks\relax
\def\floatpagepagefraction{1}
\def\textpagefraction{.001}

\shorttitle{PiPViT}    

\shortauthors{Oghbaie et al.}  

\title [mode = title]{PiPViT: Patch-based Visual Interpretable Prototypes for Retinal Image Analysis}

%
\author[1]{Marzieh Oghbaie}[type=editor,
                        orcid=0000-0001-7391-1612]
\cormark[1]
\ead{marzieh.oghbaie@meduniwien.ac.at}

\author[1]{Teresa Ara\'ujo}[orcid=0000-0001-9687-528X]
\author[1]{Hrvoje Bogunovi\'c}[orcid=0000-0002-9168-0894]
\affiliation[1]{organization={Christian Doppler Laboratory for Artificial Intelligence in Retina},
    addressline={Institute of Artificial Intelligence, Center for Medical Data Science, Medical University of Vienna}, 
    country={Austria}}

\cortext[cor1]{Corresponding author}

\begin{abstract}
Background and Objective:
Prototype-based methods improve interpretability by learning fine-grained part-prototypes; however, their visualization in the input pixel space is not always consistent with human-understandable biomarkers. In addition, well-known prototype-based approaches typically learn extremely granular prototypes that are less interpretable in medical imaging, where both the presence and extent of biomarkers and lesions are critical.

Methods:
To address these challenges, we propose PiPViT (Patch-based Visual Interpretable Prototypes), an inherently interpretable prototypical model for image recognition. Leveraging a vision transformer (ViT), PiPViT captures long-range dependencies among patches to learn robust, human-interpretable prototypes that approximate lesion extent only using image-level labels. Additionally, PiPViT benefits from contrastive learning and multi-resolution input processing, which enables effective localization of biomarkers across scales.

Results:
We evaluated PiPViT on retinal OCT image classification across four datasets, where it achieved competitive quantitative performance compared to state-of-the-art methods while delivering more meaningful explanations. Moreover, quantitative evaluation on a hold-out test set confirms that the learned prototypes are semantically and clinically relevant.
We believe PiPViT can transparently explain its decisions and assist clinicians in understanding diagnostic outcomes.
\textcolor{blue}{\hyperlink{Github page}{https://github.com/marziehoghbaie/PiPViT}}

\end{abstract}

\maketitle

\section{Introduction}
\label{sec:introduction}
Optical coherence tomography (OCT) is a crucial imaging modality in ophthalmology, offering non-invasive, high-resolution imaging of the retina. It plays a key role in diagnosing various eye conditions \cite{seebock2019exploiting, wong2014global}. An OCT scanner generates a 3D volume of the retina, composed of multiple 2D cross-sectional slices (B-scans), allowing detailed visualization of retinal layers, fovea, posterior vitreous body, and choroidal vessels. This technology supports clinicians in making informed decisions about diagnosis, treatment, and patient care \cite{jaffe2004optical}.

Advances in deep learning (DL) have enabled automated retinal disease diagnosis using architectures like convolutional neural networks (CNNs) \cite{kermany2018identifying, khan2024boost, morano2024deep} and vision transformers (ViTs) \cite{emre2023pretrained, oghbaie2024vlfatrollout}, supporting ophthalmologists in clinical applications. However, these models are often perceived as ``black boxes'' with opaque decision-making processes—a significant challenge in medical applications where incorrect predictions can have severe consequences \cite{wang2023interpretable}. 
DL models may also rely on spurious correlations in training data, known as the ``Clever Hans'' effect \cite{lapuschkin2019unmasking}, resulting in unreliable predictions in real-world scenarios. This lack of interpretability and reliance on irrelevant artifacts \cite{degrave2021ai, badgeley2019deep} has hindered the adoption of DL for classification. Addressing this black-box nature is both a legal and ethical imperative and crucial for building clinical trust and ensuring reliable outcomes in healthcare \cite{lekadir2025future}.

\begin{figure}[t]
    \centering
    \includegraphics[width=\columnwidth]{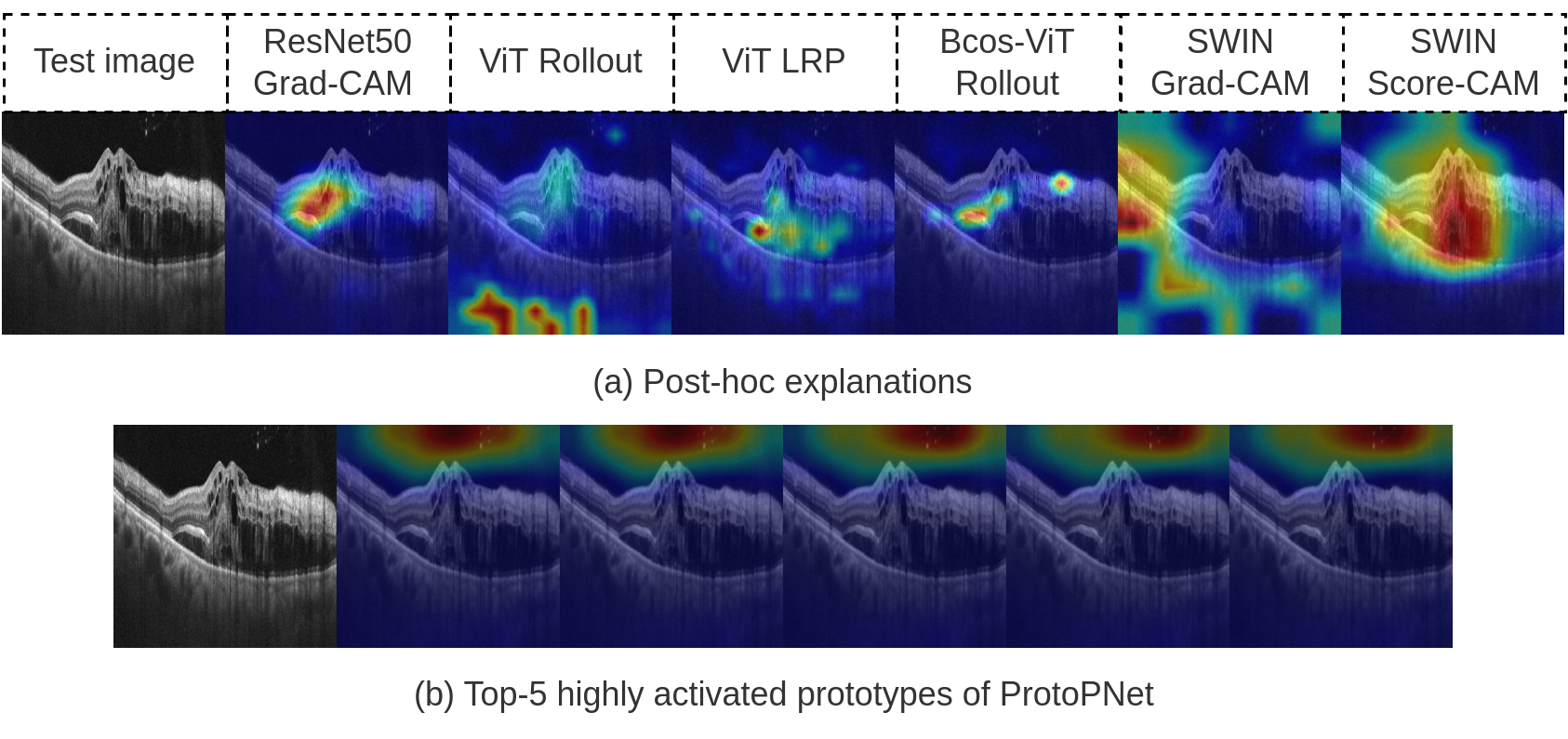}
    \caption{An RVO sample with IRF as the most prominent biomarker. (a) Post-hoc explanations for both black-box and non-prototypical approaches highlight lesions to some extent, yet they fail to fully explain the model’s decision. 
    (b) The top-5 activated prototypes of ProtoPNet are repetitive, mostly highlighting the same spatial region in the input image, which also does not correspond to IRF.}
    \label{fig:introduction}
\end{figure}

In this regard, generic post-hoc explanation methods (e.g., gradient-weighted class activation mapping (Grad-CAM) \cite{selvaraju2020grad} and Score-CAM\cite{wang2020score}) can interpret the decisions of pre-trained black-box models. However, these explanations may be insufficient in capturing fine-grained details (Fig.~\ref{fig:introduction} (a)).

\begin{figure}[t]
    \centering
    \includegraphics[width=0.9\columnwidth]{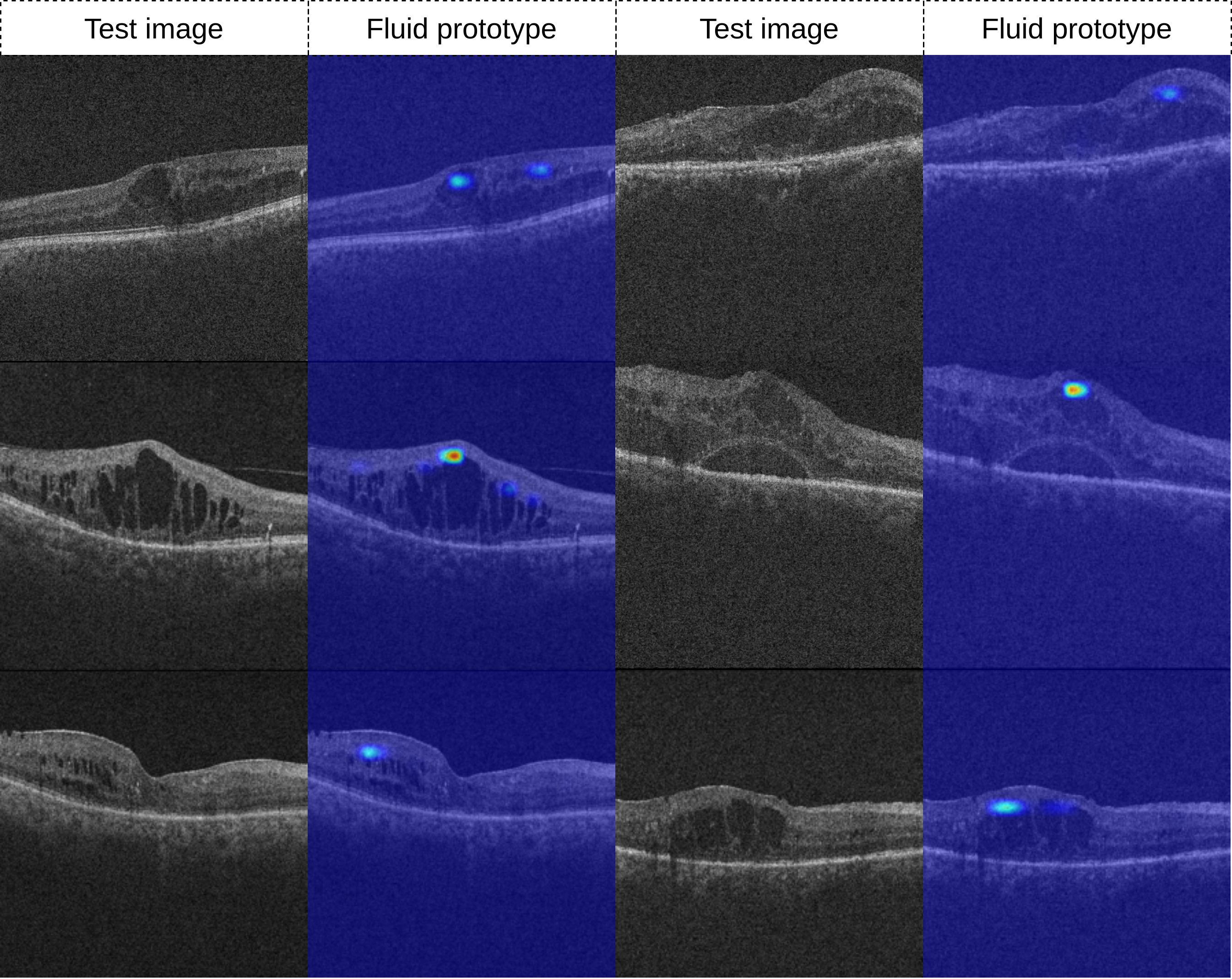}
    \caption{
    Activation map of the prototype extracted by PiP-Net corresponding to fluid in DME cases. The examples illustrate varying fluid pocket sizes, where the prototype fails to approximate the extent of the affected areas.}
    \label{fig:PiP-Net_motivation}
\end{figure}

Alternatively, other approaches aim to design models in which the learning process is more aligned with human reasoning. For example, prototype-based models, e.g., ProtoPNet \cite{ProtoPNet}), are designed to learn the most representative features for each class, with the final decision being based on the degree to which these prototypes are present. Although prototype-based models offer superior interpretability compared to post-hoc explanations, their outputs are not always consistent with human-understandable biomarkers (Fig.~\ref{fig:introduction} (b)). This discrepancy, known as the ``semantic similarity'' gap, has been addressed in other works as well \cite{nauta2023pip}. 
Furthermore, prototypical methods are mostly designed to extract local, fine-grained details—ideal for tasks like bird species identification, where the presence or absence of a learned prototype may suffice.
In medical imaging, however, such granular part-prototypes can overlook the broader context and fail to capture the extent of the affected area in the input image (see Fig.~\ref{fig:PiP-Net_motivation}). 
This issue is particularly significant in retinal image analysis, 
where retinal lesions exhibit diverse morphology, size, and location and also the same biomarker can occur in different diseases. For instance, intraretinal fluid (IRF) is a common finding in retinal vein occlusion (RVO), neovascular age-related macular degeneration (nAMD), diabetic macular edema (DME), and central serous chorioretinopathy (CSC)\cite{jee2024subretinal}.

In this paper, we introduce patch-based visual interpretable prototypes (PiPViT) for retinal image analysis. PiPViT simultaneously learns class-representative prototypes and approximates lesion extent in the pixel space using only image-level guidance. By leveraging a ViT backbone, our approach captures long-range dependencies among image patches to generate semantically meaningful prototypes.
Furthermore, we employ a multi-resolution self-supervised pre-training step, which incorporates an adjustable receptive field through the learnable positional encoding of the ViT. This pre-training enhances the model's capacity to learn discriminative and generalizable features, which are then used as prototypes. Combined with contrastive learning, this approach enables PiPViT to mitigate the risk of shortcut learning and adapt to varying input resolutions.

We evaluate PiPViT across four datasets—one private and three public—assessing both disease classification performance and interpretability. Additionally, we examine its performance in weakly supervised biomarker detection on another public dataset to evaluate the quality of the learned disease-related prototypes. The results indicate that PiPViT enhances both diagnostic performance and explainability, making it particularly well-suited for medical imaging applications where understanding lesion extent is critical.

\subsection{Related Work} \label{sec:background}

Ensuring the transparency of machine learning (ML) models is essential before they can be reliably integrated into routine clinical workflows. 
These concerns have led to the development of explainable artificial intelligence (XAI) \cite{samek2019towards}, which can be mainly divided into two categories: 1) post-hoc, model-agnostic interpretability techniques and 2) inherently interpretable model architectures.

\textit{Post-hoc model-agnostic interpretability} techniques provide explanations after the model has made its predictions, which means that they do not directly influence the model’s decision-making process. 
Consequently, the explanations may fail to fully capture the model's internal complexities and nuances \cite{yang2023survey} or may be unreliable and even misleading \cite{lapuschkin2019unmasking, rudin2019stop, salahuddin2022transparency}. 
For instance, Grad-CAM exclusively uses gradients from the final convolutional block to create a coarse localization map of key regions, even though the entire network contributes to the final prediction (Fig.~\ref{fig:introduction} (a)). Grad-CAM has been extensively applied to interpret OCT B-scan classifiers \cite{han2022classifying, haihong2023kfwc, gan2023artificial, huang2019automatic, hemalakshmi2024automated, ayhan2024interpretable}.
Score-CAM \cite{wang2020score} is another post-hoc method that identifies the most important spatial regions by observing how different feature maps influence the output score. This method also preserves the network's decision process, and it has proven effective in retinal image analysis \cite{he2023interpretable}.

In contrast, \textit{inherently interpretable models} focus on developing models that are interpretable by design, thereby enabling explanations of their internal logic and the processes leading to specific outcomes. Approaches in this category can be divided into two groups.
The first group, which we term \textit{non-prototypical} approaches, focuses on modifying the model architecture to enhance interpretability during the learning process. A notable approach in this category is the attention mechanism, which is often considered to provide insight into what a model ``pays attention to'' when making predictions \cite{bohle2024b}.
For example, Med-XAI-Net \cite{shi2021improving} and LACNN \cite{fang2019attention} used attention modules to enhance lesion awareness and overall explainability in retinal OCT classification.

Although the attention mechanism can be integrated into any CNN backbone, its inherent role in ViTs makes them more interpretable than CNNs \cite{kashefi2023explainability}. Visualizing attention weights provides an intuitive way to analyze attention patterns and identify key visual tokens. However, the choice of an effective post-hoc explanation method can significantly impact the quality of the final explanation.
For example, attention rollout \cite{abnar2020quantifying} is a class-independent method that recursively merges attention matrices from transformer blocks while neglecting contributions from other network components. Beyond attention \cite{chefer2021transformer}, on the other hand, extends rollout by incorporating layer-wise relevance propagation (LRP) scores with attention matrices.
Nonetheless, we observe that both rollout and beyond attention, which we refer to as LRP, often highlight less relevant tokens, leading to noisy attention maps (see Fig.~\ref{fig:introduction} (a)).

Besides attention, B-cos transformation \cite{bohle2024b} is another method that aims to bring explainability into the model architecture. It is designed such that its weights align with task-relevant input signals during optimization. However, this approach has been shown to be data-intensive and requires long training procedures with more augmentations.

The second approach for developing inherently interpretable models is \textit{prototype learning}, a well-established method for constructing transparent models that align with human reasoning. Among prototypical networks, ProtoPNet \cite{ProtoPNet} is widely used and explains new examples by comparing image parts with learned prototypes. ProtoPNet has been applied in various medical imaging modalities, including brain MRI \cite{mohammadjafari2021using}, chest CT \cite{singh2021interpretable,singh2022think}, whole-slide images \cite{ProtoMIL}, and retinal OCT \cite{djoumessi2024actually}.

However, prototypical networks often trail black-box models in accuracy, with some efforts using teacher-student approaches to boost performance \cite{wang2023interpretable}. The semantic gap between the learned prototypes and meaningful input concepts also limit the explainability of prototypical networks. While Barnett et al. \cite{barnett2021case} partially addressed this with extra pixel-level annotations, the issue can stem from fixing the number of prototypes per class: too many result in redundancy, while too few force merging of distinct concepts (see Fig.~\ref{fig:introduction} (b)). 

In contrast to traditional prototype-based approaches, Nauta et al. \cite{nauta2023pip} introduced PiP-Net, which does not explicitly learn prototype embeddings. Instead, the features extracted by the network are directly treated as prototypes, allowing PiP-Net to impose only an upper bound on the total number of prototypes. The model then learns a sparse classification layer that selects the most salient prototypes for each class by applying sparsity regularization. Although PiP-Net improves interpretability and performance in both natural and medical imaging, it remains prone to shortcut learning by relying on clinically irrelevant artifacts \cite{nauta2023interpreting}. 

In general, while post-hoc explanations for black-box models are useful for understanding model results, they do not provide insight into the internal logic of the models. On the other hand, non-prototypical inherently interpretable approaches aim to modify the model's learning process, but their explanations still cannot fully reveal their internal mechanisms. Prototypical models have proven to follow a more transparent, step-by-step decision-making process; however, the quality of prototypes and their corresponding interpretation in the input space can be limited due to locality and the fixed-number-of-prototypes-per-class constraints.
Our PiPViT model, inspired by PiP-Net, also treats extracted features as prototypes and employs sparsity regularization. Additionally, it incorporates image-level attention and multi-resolution pre-training to effectively identify prototypical lesions and approximate their spatial extent.

\section{Methods}
\subsection{Overview}
\begin{figure*}[t]
    \centering
    \includegraphics[width=\textwidth]{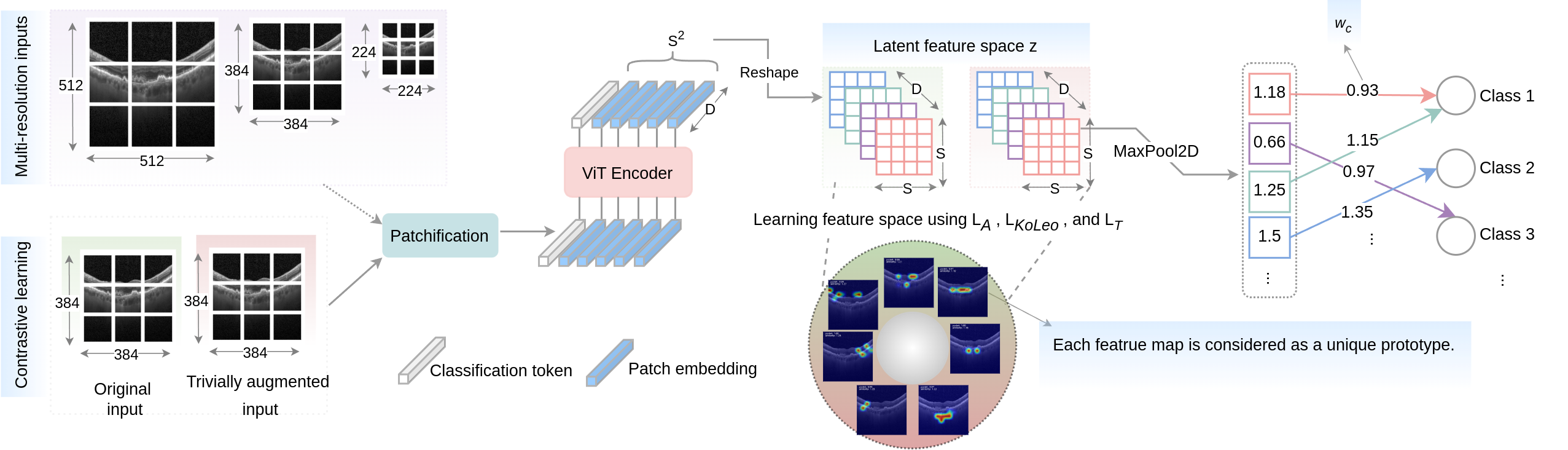}
    \caption{
    The pipeline of our proposed PiPViT: The ViT encoder extracts patch representations from the input image at a given resolution. Pre-training is conducted using three different image resolutions, with consistent patch size and adaptively resized positional embeddings. The sequence of patches is reshaped into \( S \times S \times D \) feature maps, where the pooled value of each feature map represents presence scores.    
    Contrastive learning, guided by alignment (\( L_A \)), tanh-loss \( L_T \), and KoLeo (\( L_{KoLeo} \)) losses, clusters similar features that might represent a single biomarker together in the latent space. Tanh-loss prevents trivial solutions and feature collapse. Finally, a sparse linear layer connects learned part-prototypes to classes, making the model’s output interpretable as a scoring sheet.}
    \label{fig:main-arch}
\end{figure*}

An overview of the proposed approach is illustrated in Fig.~\ref{fig:main-arch}. Given an input image \(x \in \mathbb{R}^{W \times H}\), where \(W\) and \(H\) represent width and height, PiPViT utilizes a ViT backbone to extract low-level feature maps \(z \in \mathbb{R}^{W' \times H' \times D}\). Here, \(W'\) and \(H'\) (denoted as \(S\) in Fig.~\ref{fig:main-arch}) correspond to the spatial dimensions of the feature map, while \(D\) represents the number of features.
More specifically, each feature map with shape \(W' \times H'\) is considered as a prototype and \(D\) is the upper bound for the total number of prototypes that the model learns. 
To improve the semantic quality of the feature space \( \mathcal{F} \) and ensure resolution-agnostic feature learning, we pre-trained the PiPViT with a multi-resolution contrastive learning approach (see Section \ref{sec:featrue_learning}).

Following \cite{nauta2023pip}, a softmax operation is applied over \(D\) to encourage each patch \(z_{w', h', :}\) to strongly associate with a single prototype, making the distribution more concentrated.
The max-pooled value of each feature map \(z_{:, :, d}\) quantifies the activation strength of prototype \(d\) in the input image \(x\), resulting in a vector \(p \in [0, 1]^D\).
Vector \(p\) is then fed to a fully connected (FC) layer with sparsity regularization to learn the association between each prototype and a given class (see Section~\ref{sec:sparsity}). This regularization enhances interpretability by enforcing non-negativity on the weights, leading to a sparse scoring sheet.

\subsection{Forming a Discriminative Feature Space}
\label{sec:featrue_learning}

\paragraph{Feature-level Objectives}
First, we optimize the feature space with alignment (\(L_A\)) and tanh (\(L_T\)) losses \cite{nauta2023pip}. 
\(L_A\) optimizes the feature space by ensuring that two views of the same image patch correspond to the same, ideally, single prototype. Consequently, this makes the feature space invariant to irrelevant noise factors.
\begin{equation}
{L}_A = - \frac{1}{W'H'} \sum_{(w', h') \in W' \times H'} \log(z'_{w', h',:} \cdot z''_{w', h',:}).
\end{equation}

\(L_T\) prevents trivial solutions and ensures prototype diversity within each mini-batch. 
The tanh loss term corresponds to:

\begin{equation}
    {L}_T(p) = -\frac{1}{D} \sum_{d}^{D} \log(\tanh(\sum_{b}^{B} (p_b)) + \epsilon),
\end{equation}

where $\tanh$ and $\log$ are element-wise, \(B\) is the number of samples in a mini-batch and \(\epsilon\) is a small number for numerical stability. 

Moreover, to introduce both locality and discrimination in the feature space while preserving maximal information, we incorporate KoLeo regularization (\({L}_{KoLeo}\)) \cite{oquab2023dinov2}, based on the Kozachenko-Leonenko estimator. 
Previously applied to ViT's feature space \cite{oquab2023dinov2}, KoLeo encourages a uniform feature span within a batch using a differentiable nearest-neighbor density estimation, computed in \(O(n \log n)\) \cite{sablayrolles2018spreading}. Geometrically, it pushes the closest points apart with a concave, non-decreasing force to prevent excessive clustering and ensure broad coverage of the hypersphere for robust and distinctive prototype learning.
Given a set of \(n\) feature maps in a batch \((z_1, ..., z_n)\), the KoLeo loss term corresponds to:

\begin{equation}
{L}_{KoLeo} = - \frac{1}{n} \sum_{i=1}^{n}\log({\rho}_{n , i})
\end{equation}

where \(\rho_{n , i} = min_{j \neq i} \norm {{z}_{i} - {z}_{j}}\) is the minimum distance between feature map \(i\) and any other feature map \(j\) within the batch.
Finally, the pre-training objectives for the ViT backbone is as follows:

\begin{equation}\label{eq:pre-training}
    {L}_{pre-train} = \lambda_A {L}_A + \lambda_T {L}_T + \lambda_{KoLeo} {L}_{KoLeo}
\end{equation}

\paragraph{Multi-Resolution pre-training} 
Input resolution is particularly crucial for medical image processing for two main reasons:
first, in real-world applications, test images may originate from different capturing devices than those used during development, leading to resolution variations and potential ``out-of-domain'' issues. This lack of generalization to unseen resolutions can compromise prediction reliability.
Second, lower resolutions impede the detection of small biomarkers, while higher resolutions enhance feature quality at the cost of increased computational demands. Two practical approaches to address this challenge include temporarily increasing the resolution for a limited number of training epochs \cite{likhomanenko2021cape, oquab2023dinov2} or varying the patch size during development \cite{beyer2023flexivit}.

Building on these ideas, we introduce a multi-resolution pre-training step for the ViT backbone. This method pre-trains the model across a predefined set of input resolutions while the patch size remains fixed. This configuration balances computational efficiency and prototype granularity. For a new resolution, the only component of the ViT that requires modification is the positional embedding sequence, as different input resolutions yield a varying number of visual tokens.
After this adjustment, the model is pre-trained using the objective specified in Eq. \ref{eq:pre-training} for a predetermined number of epochs. This process is repeated for all provided resolutions, resulting in a model that is progressively pre-trained at multiple resolutions.
Since the extent and location of lesions can vary between samples within a class, this multi-resolution pre-training can mitigate spurious correlations between content and position, thereby enhancing the fine-grained detection of lesions of varying sizes.

\subsection{Interpretable Classifier}
\label{sec:sparsity}
After pre-training, the entire network is fine-tuned to learn \(w_c \in \mathbb{R}^{D \times K}_{\geq 0} \), which determines the contribution of each prototype to the final prediction.  
In particular, a sparse linear layer is optimized using the classification loss (\({L}_C\)), defined as the negative log-likelihood loss between the predicted label \(\bar{y}\) and the one-hot encoded ground truth \(y\). This loss is further combined with self-supervised objectives:

\begin{equation}\label{eq:finetuning}
{L} = \lambda_A {L}_A + \lambda_T {L}_T + \lambda_{KoLeo} {L}_{KoLeo} + \lambda_C {L}_C
\end{equation}

The sparsity is enforced using the \(\log((pw_c)^n + 1)\) regularizer where \(p \in [0, 1]^D\) is the prototype presence score, \(w_c\) the linear layer weight, and \(n\) the regularization order. Although \(n\) is set to two by default \cite{nauta2023pip}, we suggest higher values to enhance the sensitivity to strongly activated prototypes while reducing the influence of weaker ones.  

\section{Evaluation}\label{sec:evaluation}

We evaluated the applicability and potential of PiPViT for the classification of retinal images by analyzing its classification performance (Section \ref{sec:classification_results}), followed by the qualitative evaluation of the learned prototypes (Section \ref{sec:explainability}). Moreover, to ensure prototypes' relevance in medical contexts, in Section \ref{sec:drusen_detection}, we assessed whether they align with human reasoning and correspond to known biomarkers.

\subsection{Datasets}\label{sec:datasets} 

We conducted experiments using five retinal OCT datasets (four public and one private) for the B-scan classification task. 

\begin{itemize}
    \item The \textbf{Kermany} dataset \cite{kermany2018identifying} consists of $108309$ OCT images from $4686$ patients for training and validation and $1000$ images from $633$ patients for testing. It includes four classes: choroidal neovascularization (CNV), DME, drusen, and Normal. All images were acquired using the Spectralis OCT system (Heidelberg Engineering, Germany). To ensure consistency with prior work, we used the official train-test split and allocated $20$\% of the training set patients for validation.   

    \item The OCT image database (\textbf{OCTID}) \cite{gholami2020octid} consists of $572$ OCT images\footnote{The number of patients is not officially available, thus it is assumed that each sample is from a different patient.}, encompassing five categories: Normal, macular hole (MH), age-related macular degeneration (AMD), central serous retinopathy (CSR), and diabetic retinopathy (DR). The images were acquired using the Cirrus HD-OCT machine (Carl Zeiss Meditec, Inc., Dublin, CA).
    The data were split by patients into $50$\% for training, $25$\% validation, and $25$\% for the test. 
    
    \item The OCT dataset for image-based deep learning (\textbf{OCTDL}) \cite{kulyabin2024octdl} contains $2064$ images from $821$ patients, categorized into seven different eye conditions: AMD, DME, epiretinal membrane (EM), Normal, retinal artery occlusion (RAO), RVO, and vitreomacular interface disease (VID). The images were captured using the Optovue Avanti RTVue XR system.
    The data splits are made similar to OCTID.
    
    \item \textbf{OCT5K} \cite{arikan2023oct5k} is a subset of the Kermany dataset with B-scan-level biomarker annotations, including drusen bounding boxes. This dataset was used to evaluate the semantic quality of the learned prototypes. For our experiments, we utilized $105$ B-scans containing $1132$ annotated drusen boxes.
    
    \item Private \textbf{OPTIMA5C} dataset \cite{oghbaie2024vlfatrollout} includes $4416$ treatment-naive OCT volumes from $3739$ patients, collected from baseline scans of multicenter clinical trials available at the Department of Ophthalmology, Medical University of Vienna. It covers five disease categories:  intermediate age-related macular degeneration (iAMD), nAMD, geographic atrophy (GA), DME, and Normal, with scans captured by Cirrus (Zeiss Meditec) and Spectralis (Heidelberg Engineering) devices. In this paper, only central B-scans were used, as the foveal region is critical for macular disease diagnosis. Notably, the test and development (train and validation) splits are determined not only on a patient-wise basis but also according to clinical studies, rendering OPTIMA5C more challenging than previously mentioned datasets. The dataset is split into training ($2372$ samples, $2041$ patients), validation ($591$ samples, $511$ patients), and test ($1453$ samples, $1187$ patients) sets. 
\end{itemize}

\subsection{Baselines}
\label{sec:baselines}
To evaluate the classification performance of PiPViT on the retinal B-scan classification task, we conducted a comparative analysis using models with varying levels of explainability, ranging from black-box models to prototype-based approaches.

\paragraph{Black-box and Non-Prototypical Models}
As a black-box baseline, we employed ResNet-50 \cite{he2016deep} due to its established effectiveness in retinal OCT analysis \cite{rasti2017macular, emre2023pretrained}.
Among non-prototypical approaches, we included ViT and shifted windows transformer (SWIN) \cite{liu2021swin} as attention-based baselines. We also incorporated B-cos ViT \cite{bohle2024b} which has demonstrated promising performance in image classification.

\paragraph{Prototype Learning}
We compared PiPViT with both PiP-Net \cite{nauta2023pip} and ProtoPNet \cite{ProtoPNet}. As an additional baseline, we also applied the prototype pruning method, ProtoPShare \cite{rymarczyk2020protopshare}. In this data-dependent merge-pruning mechanism, a specific percentage of most similar prototypes are merged\footnote{For our experiments, we used the default value which is set to $0.1$.}. ProtoTrees \cite{nauta2021neural} is another baseline consisting of a CNN followed by a binary tree structure.
We also included ProtoPFormer \cite{xue2022ProtoPFormer}, a variation of ProtoPNet leveraging ViT for feature extraction and attention rollout to enhance the model’s focus on foreground details. 

\paragraph{Evaluation of Classification Performance}
To assess the classification performance, we reported balanced accuracy (BAcc), area-under-the-curve (AUC), and F1-score. We used bootstrapping with $1000$ replicates on the hold-out test sets to estimate the variance in model performance.

\paragraph{Explainability Techniques}
For black-box and non-prototypical approaches, we deploy a post-hoc method to explain the decisions. However, given the differences in model mechanisms and challenges in reproducibility, no single explainability method works universally. For example, Grad-CAM produced reasonable maps for ResNet-50 but failed on the SWIN transformer (see Fig.~\ref{fig:introduction} (a)).
As a result, we selected explainability methods validated in prior studies. Specifically, following \cite{han2022classifying} and \cite{he2023interpretable}, we applied Grad-CAM to ResNet-50 and Score-CAM to the SWIN transformer, respectively. For the ViT baseline and B-cos-ViT, we visualized attention rollout\cite{oghbaie2024vlfatrollout}. Additionally, we incorporated LRP applied to the ViT baseline \cite{komorowski2023towards, zhou2023foundation}.

In contrast, prototypical models avoid the challenge of selecting a visualization method, as prototypes can be directly visualized in the input space for more intuitive interpretability. 

\subsection{Evaluation of Semantic Alignment of Prototypes with Biomarkers}\label{sec:proto_eval}
\begin{figure}[t]
    \centering
    \includegraphics[width=0.9\columnwidth]{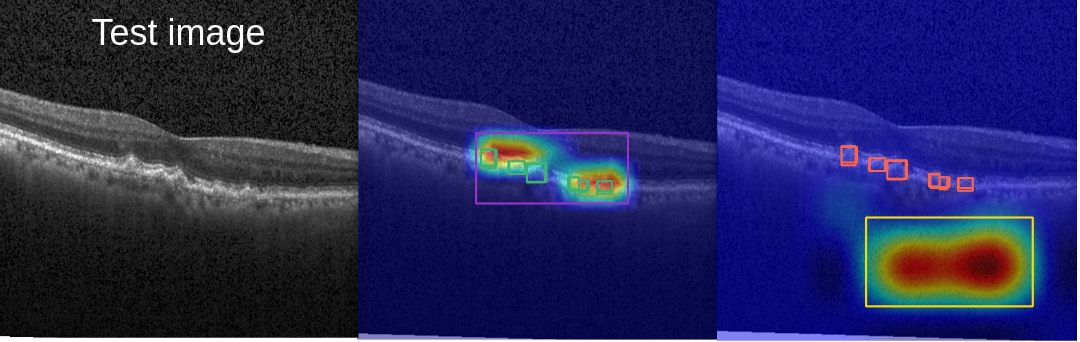}
    \caption{Drusen detection is framed as an object detection task by mapping the drusen prototype into pixel space to highlight potential drusen regions (purple box). A true positive occurs when the ground truth overlaps the highlighted region (green); a false negative when a drusen region is missed (red); and a false positive when a non-drusen area is misclassified as drusen (yellow).}
    \label{fig:biomarker_detection}
\end{figure}

In this experiment, we evaluated the semantic quality of the learned prototypes to ensure they align with known biomarkers and provide relevant information in a medical context. To do so, we framed the evaluation of drusen, yellow deposits composed of lipids and proteins beneath the retina \cite{garcia2017early}, prototypes as an object detection task, where a prototype should activate over multiple drusen regions.
More specifically, we first trained a binary classifier on a balanced subset\footnote{List of samples in each split are available in our project repository.} of Kermany dataset containing only drusen and Normal classes. We then used the OCT5K drusen bounding box annotations to assess the semantic quality of the drusen prototypes.
In this setting, ground truth drusen within the activation map are considered true positives, undetected drusen as false negatives, and healthy regions detected as drusen as false positives. True negatives are not defined in this setting (see Fig.~\ref{fig:biomarker_detection}).

To generate potential bounding boxes from the learned prototypes, we placed a box around each highlighted region in the activation map as a potential drusen region (see Fig.~\ref{fig:biomarker_detection}). The box size was controlled by a scale factor ranging between \([0.2, 10]\) in increments of $0.1$, and precision and recall were computed for each scale factor. Finally, we calculated average precision (AP) \cite{marimont2023harder} for insights across different scales.

\subsubsection{Sensitivity Analysis of PiPViT's Pre-training on the Semantic Quality of Prototypes}
In this subsection, we study the semantic alignment between the learned prototype and the actual biomarker in various pre-training configurations. First, we examine the impact of input and patch sizes, with PiPViT pre-trained using the single-resolution strategy. Next, we explore the effect of different pre-training strategies, comparing no pre-training with single- and multi-resolution pre-training.
Note that these experiments specifically focus on drusen detection, as it provides an ideal scenario for evaluating prototype semantic quality and understanding the extent of the affected area under different conditions.

\subsection{Experimental Settings}
All experiments were conducted using PyTorch 1.13.0+cu117 and timm 0.5.4 \cite{rw2019timm} on a server equipped with 1TB RAM and an NVIDIA RTX A6000 (48GB VRAM).
All models were trained for $500$ epochs with a batch size of 64, the best model was selected based on BAcc on the validation set, and the final performance was reported on a hold-out test set.
We used AdamW \cite{loshchilov2017decoupled} with cosine annealing and warmup, and tuned learning rates via grid search. 
All the models were initialized with pre-trained ImageNet weights \cite{russakovsky2015imagenet} and processed \(224 \times 224\) grayscale inputs ($3$ channels\footnote{To utilize the pre-trained weights, we replicate the grayscale slice across three channels to form a three-channel input image.}). ViT-based models used a patch size of $16$.  During training, random augmentations \cite{cubuk2020randaugment} were applied, while validation and test images were only resized and normalized. 
For classification, we used focal loss \cite{lin2017focal} to emphasize hard, misclassified examples. 

Both PiPViT and PiP-Net were pre-trained separately on the training set without using labels. For PiPViT, the input size and patch size were set to $384$ and $32$, respectively. During multi-resolution pre-training, the model was trained for $10$ epochs at each resolution ($224$, $384$, and $512$) using a batch size of $64$, and the model with the lowest \({L}_{pre-train}\) was selected for fine-tuning.
For PiP-Net, the pre-training phase lasted $100$ epochs with a batch size of $128$. For contrastive pre-training of both PiPViT and PiP-Net, we followed \cite{nauta2023pip} and employed TrivialAugment \cite{muller2021trivialaugment}. In both models, we used negative log-likelihood loss.

Regarding the hyperparameters (\(\lambda_A\), \(\lambda_T\), \(\lambda_C\), \(\lambda_{KoLeo}\)) in the pertaining and fine-tuning loss terms of PiPViT (see Eq. \ref{eq:pre-training} in Section \ref{sec:featrue_learning} and Eq. \ref{eq:finetuning} in Section \ref{sec:sparsity}), as well as \(n\) (see Section \ref{sec:sparsity}), we selected reasonable values without performing an extensive hyperparameter search to minimize carbon footprint. 
During pertaining, \(\lambda_A\) was gradually increased to $1$ as a warm start to prevent the trivial solutions.
A similar strategy was adopted for setting the hyperparameters of the loss functions in PiP-Net.

\section{Results and Discussion}
\subsection{Classification Results}\label{sec:classification_results}

\begin{figure}[]
\begin{subfigure}{0.5\textwidth}
\includegraphics[width=0.95\linewidth]{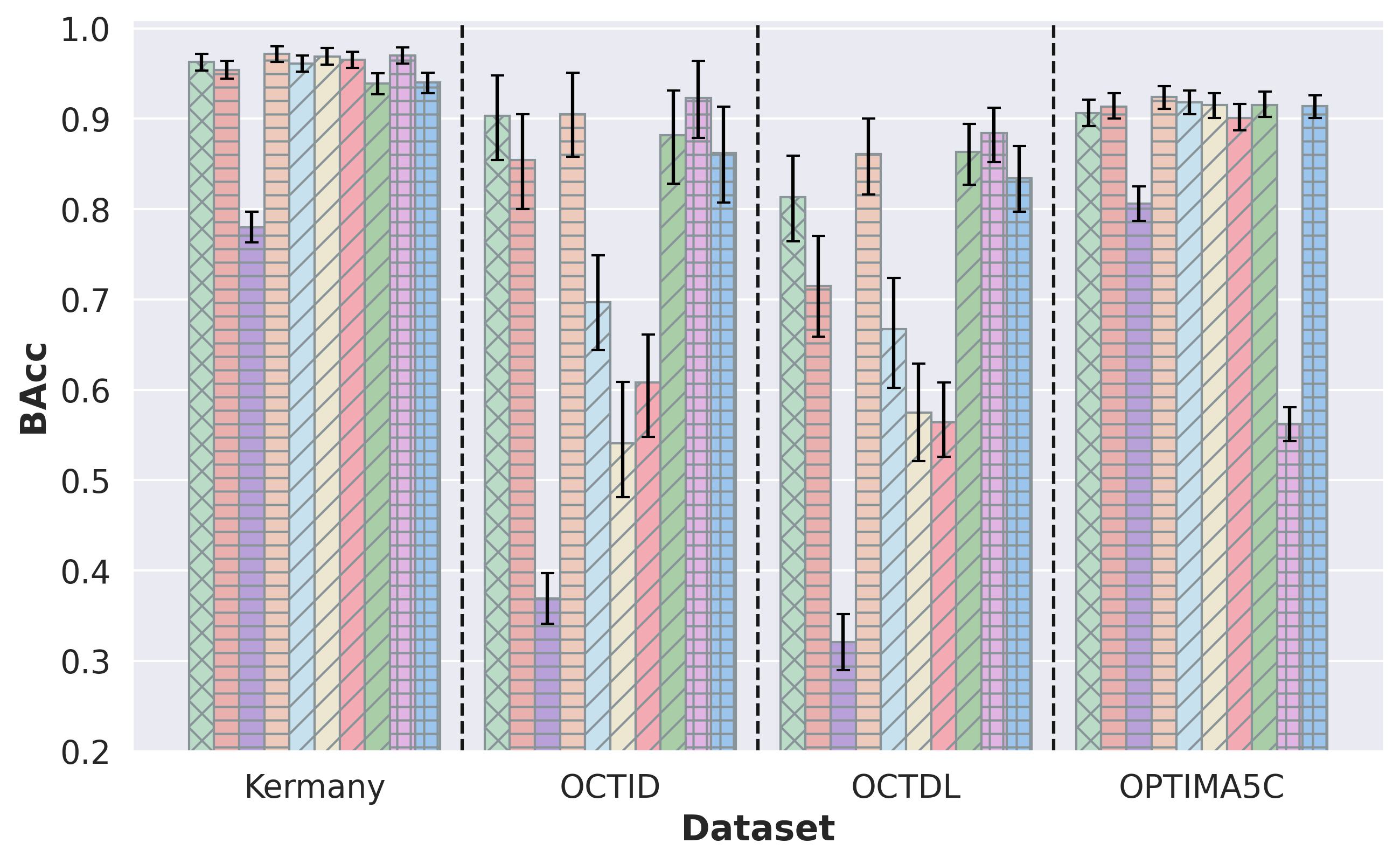} 
\end{subfigure}
\begin{subfigure}{0.5\textwidth}
\includegraphics[width=0.95\linewidth]{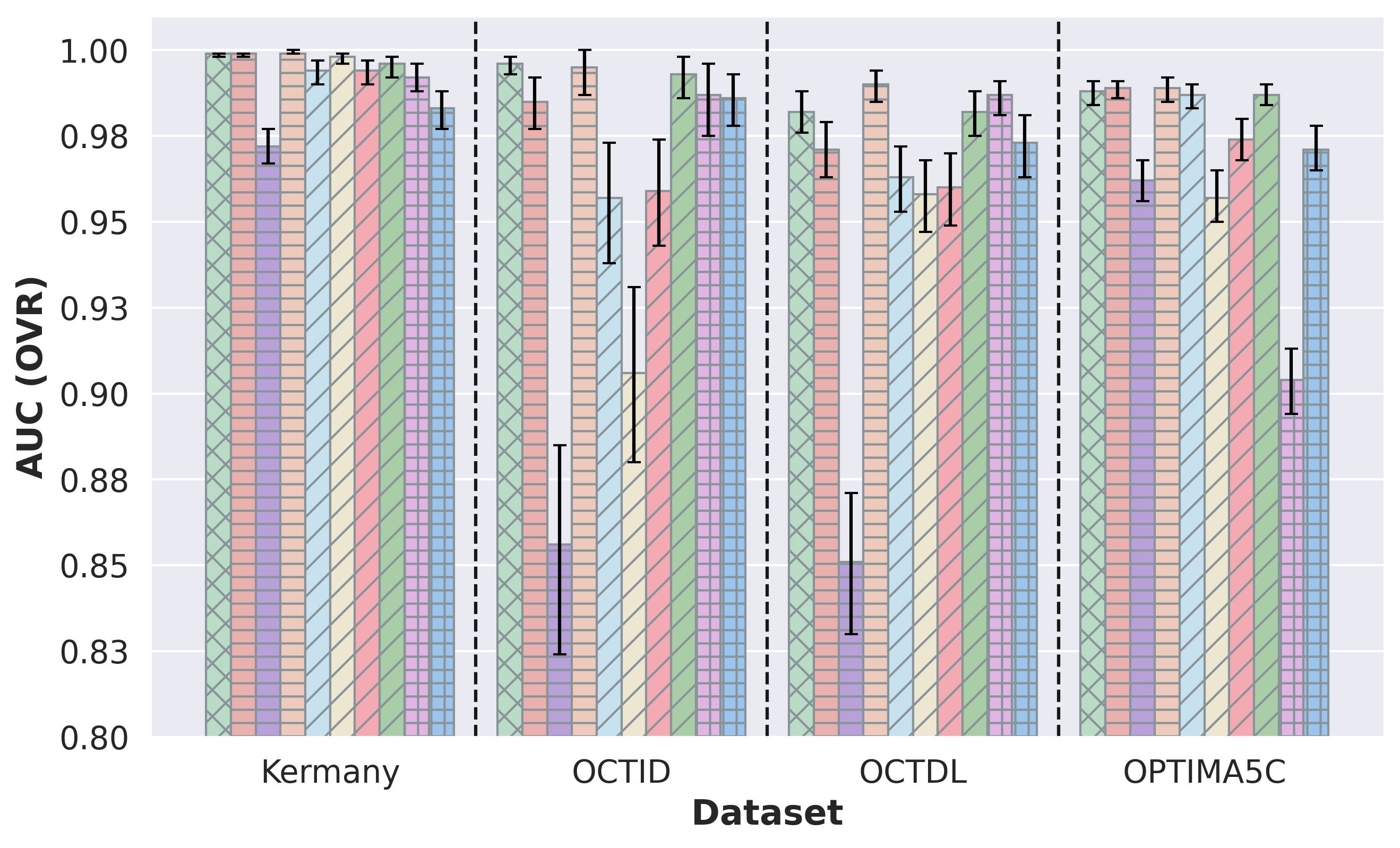} 
\end{subfigure}
\begin{subfigure}{0.5\textwidth}
\includegraphics[width=0.95\linewidth]{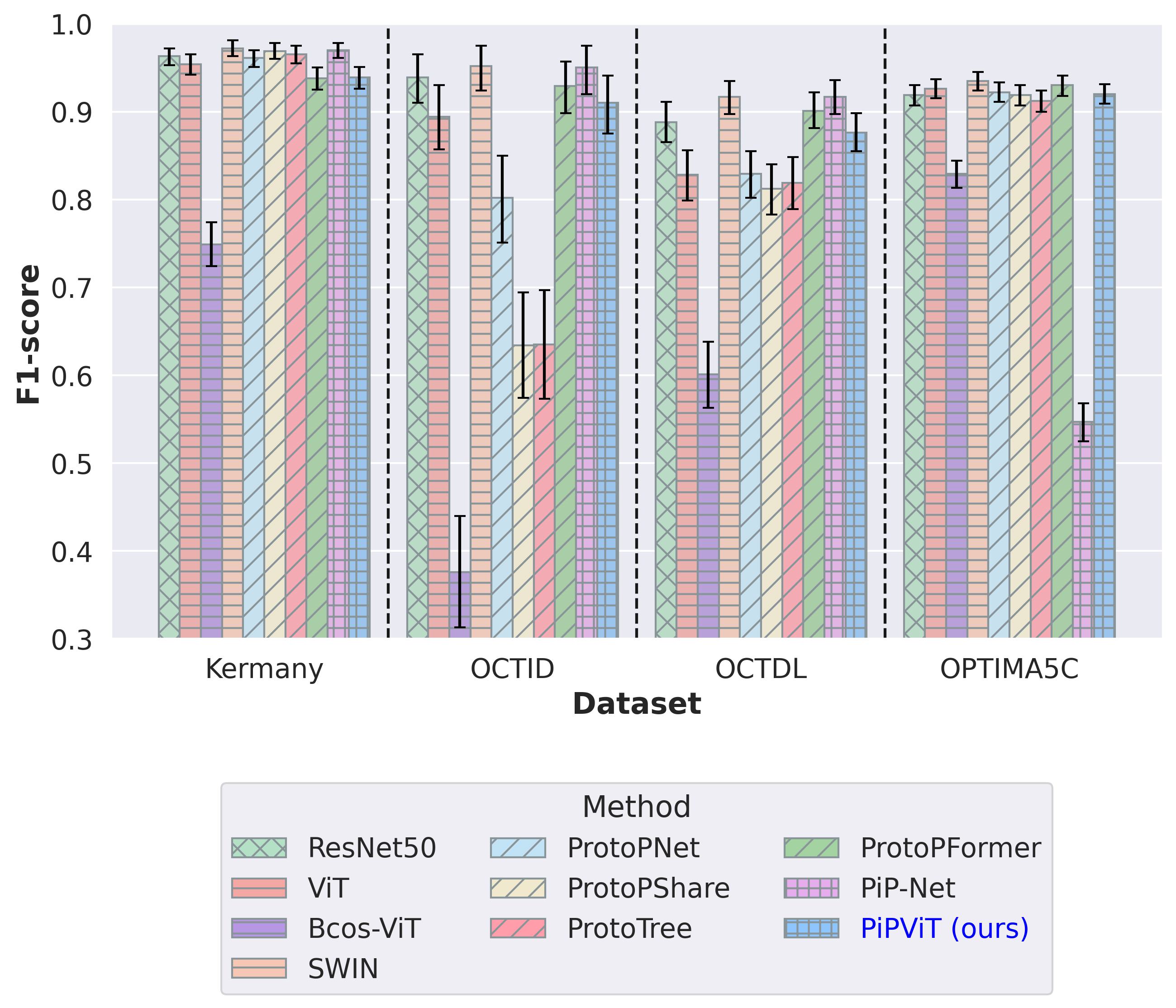} 
\end{subfigure}
\caption{Classification results on the test set, with $95$\% confidence intervals from $1000$ bootstrap replicates shown as error bars. To facilitate comparison between methods, different patterns are used for each subgroup: black-box models are represented with ``xx'', non-prototypical models with ``--'', traditional prototype-based models with ``//'', and models that directly use features as prototypes with ``++''.}
\label{fig:ClassificationPerformance}
\end{figure}

\begin{table*}[]
\caption{Average classification performance \(\pm\) std. of baselines and PiPViT over all the datasets.}
\label{tab:overall}
\resizebox{\textwidth}{!}{%
\begin{tabular}{@{}lllllllllll@{}}
\toprule
\textbf{Metric} &
  \multicolumn{1}{c}{\textbf{ResNet50}} &
  \multicolumn{1}{c}{\textbf{ViT}} &
  \multicolumn{1}{c}{\textbf{Bcos-ViT}} &
  \multicolumn{1}{c}{\textbf{SWIN}} &
  \multicolumn{1}{c}{\textbf{ProtoPNet}} &
  \multicolumn{1}{c}{\textbf{ProtoPshare}} &
  \multicolumn{1}{c}{\textbf{ProtoTree}} &
  \multicolumn{1}{c}{\textbf{ProtopFormer}} &
  \multicolumn{1}{c}{\textbf{PiPNet}} &
  \multicolumn{1}{c}{\textbf{\textcolor{blue}{PiPViT}}} \\ \midrule
\textbf{AUC (OVR)} &
  0.991 ± 0.008 &
  0.986 ± 0.012 &
  0.910 ± 0.066 &
  0.993 ± 0.005 &
  0.975 ± 0.018 &
  0.955 ± 0.038 &
  0.972 ± 0.016 &
  0.990 ± 0.006 &
  0.968 ± 0.042 &
  0.978 ± 0.007 \\
\textbf{F1-Score} &
  0.927 ± 0.032 &
  0.901 ± 0.054 &
  0.639 ± 0.199 &
  0.944 ± 0.024 &
  0.879 ± 0.075 &
  0.834 ± 0.148 &
  0.833 ± 0.145 &
  0.925 ± 0.016 &
  0.846 ± 0.201 &
  0.911 ± 0.026 \\
\textbf{BAcc} &
  0.896 ± 0.062 &
  0.859 ± 0.104 &
  0.569 ± 0.260 &
  0.916 ± 0.046 &
  0.811 ± 0.150 &
  0.750 ± 0.223 &
  0.760 ± 0.203 &
  0.900 ± 0.034 &
  0.835 ± 0.185 &
  0.888 ± 0.048 \\ \midrule
\textbf{Model size (MB)} &
  \multicolumn{1}{c}{89.920} &
  \multicolumn{1}{c}{327.311} &
  \multicolumn{1}{c}{326.339} &
  \multicolumn{1}{c}{332.422} &
  \multicolumn{1}{c}{122.716} &
  \multicolumn{1}{c}{122.278} &
  \multicolumn{1}{c}{92.389} &
  \multicolumn{1}{c}{327.918} &
  \multicolumn{1}{c}{89.920} &
  \multicolumn{1}{c}{333.921} \\ \bottomrule
\end{tabular}%
}
\end{table*}

Table ~\ref{tab:overall} and Fig. \ref{fig:ClassificationPerformance} summarize the classification performance of PiPViT and of the tested baselines across four retinal OCT datasets.
Results show that the proposed method, PiPViT, achieved on-par performance with both prototypical and non-prototypical baselines, demonstrating the effectiveness of the proposed pipeline for learning class-representative prototypes. 
The strong classification performance of PiPViT and PiP-Net highlighted the benefits of using features as prototypes over traditional prototype learning (e.g., ProtoPNet, ProtoTree). Notably, on the OCTID and OCTDL datasets, ProtoPNet, ProtoPshare, and ProtoTree exhibited the lowest performance among prototypical networks.

In general, PiP-Net achieved competitive classification performance on the public datasets; however, its performance dropped significantly on OPTIMA5C, which includes more challenging cases with varying disease severity levels. We believe this decline was due to the excessive granularity of PiP-Net’s feature maps, which prevented them from fully capturing the affected regions, thus limiting its ability to handle the high intra-class variability in OPTIMA5C (see Fig.~\ref{fig:PiP-Net_motivation}).
On the other hand, by leveraging attention to extract relationships among patches, PiPViT captured better the global context of the B-scan and, consequently, the extent of the lesion. This also allowed PiPViT to extract features that generalize more effectively to the OPTIMA5C test samples, which come from different studies and may contain characteristics not seen during training.

The average classification performance (Table \ref{tab:overall}) among prototypical networks revealed that models with a ViT backbone— PiPViT and ProtoPFormer—consistently outperformed those with a CNN-based backbone. 
For example, ProtoPFormer achieved a $10.97$\% average improvement in BAcc over its CNN-based counterpart, ProtoPNet. Overall, ProtoPFormer achieved higher BAcc, although the difference compared to PiPViT was small (only a $1.3$\% decrease for PiPViT).

Overall, the SWIN Transformer achieved the best classification performance across all datasets, on average $3.1$\% better than PiPViT. This superior performance is largely due to its hierarchical design, enabling it to generalize more effectively across diverse datasets.
We note that B-cos-ViT demonstrated the lowest performance among baselines. This performance decrease may be attributed to the fact that B-cos models benefit from a longer training protocol with additional data enhancement \cite{bohle2024b}. However, for fairness, we used the same training procedure and augmentation across all baselines.

\subsection{Evaluation of Model Explainability} \label{sec:explainability}

\begin{figure}[t]
    \centering
    \includegraphics[width=\columnwidth]{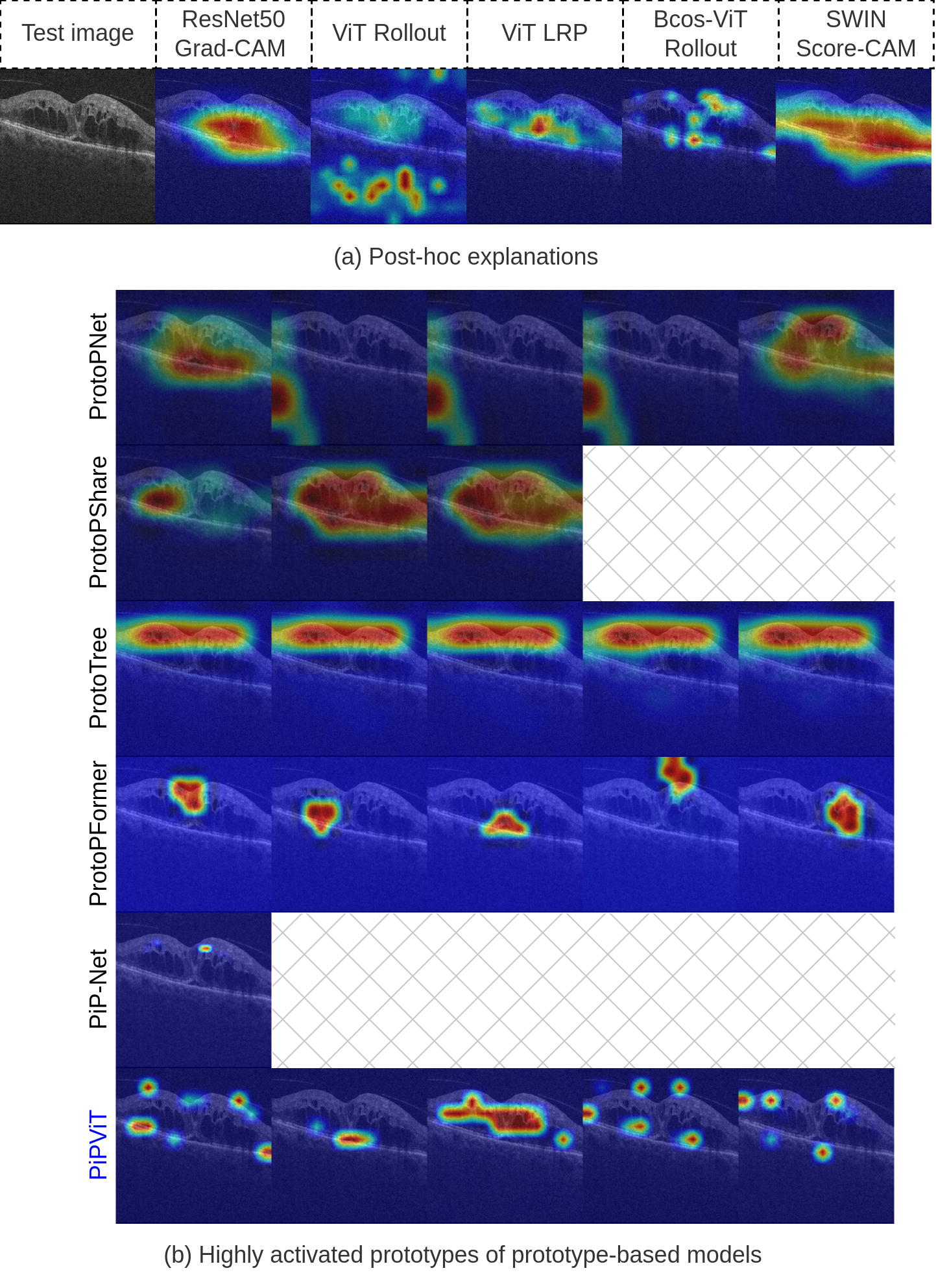}
    \caption{Visual explanation of different models' decisions for a DME sample in the OPTIMA5C.
    (a) Heatmaps generated by post-hoc approaches applied to black-box and non-prototypical models.
    (b) Top-5 highly activated prototypes of PiPViT vs. baseline prototypical methods (ProtoPShare and PiP-Net had less than five prototypes). While traditional prototype-based models struggled to capture key DME features, the first three prototypes of PiPViT preserved feature diversity and effectively identified critical biomarkers, including a swollen retina, RPE alterations, and intra-retinal fluid.}
    \label{fig:dme_heatmaps}
\end{figure}

\begin{figure}[t]
    \centering
    \includegraphics[width=\columnwidth]{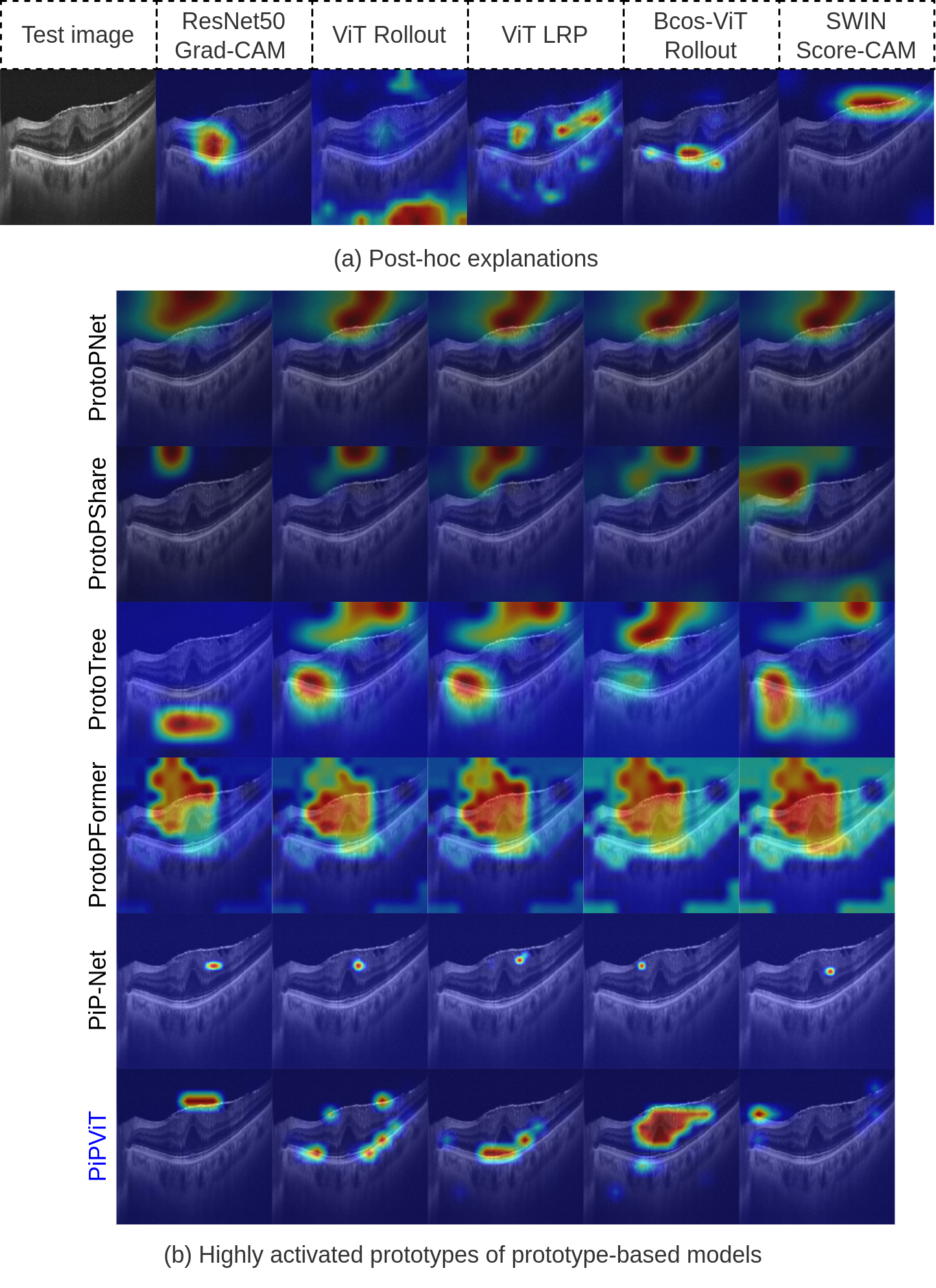}
    \caption{Visual explanation of different models' decisions for an ERM sample in OCTDL.
    (a) Heatmaps generated by post-hoc approaches applied to black-box and non-prototypical models.
    (b) Top-5 highly activated prototypes of PiPViT vs. baseline prototypical methods: PiPViT identified critical features of ERM, including epiretinal fibrosis, and increased central foveal thickness (represented by the first and fourth prototypes, respectively), demonstrating superior alignment with pathology-relevant biomarkers.}
    \label{fig:erm_heatmaps}
\end{figure}

\begin{figure}[]
    \centering
    \includegraphics[width=0.9\columnwidth]{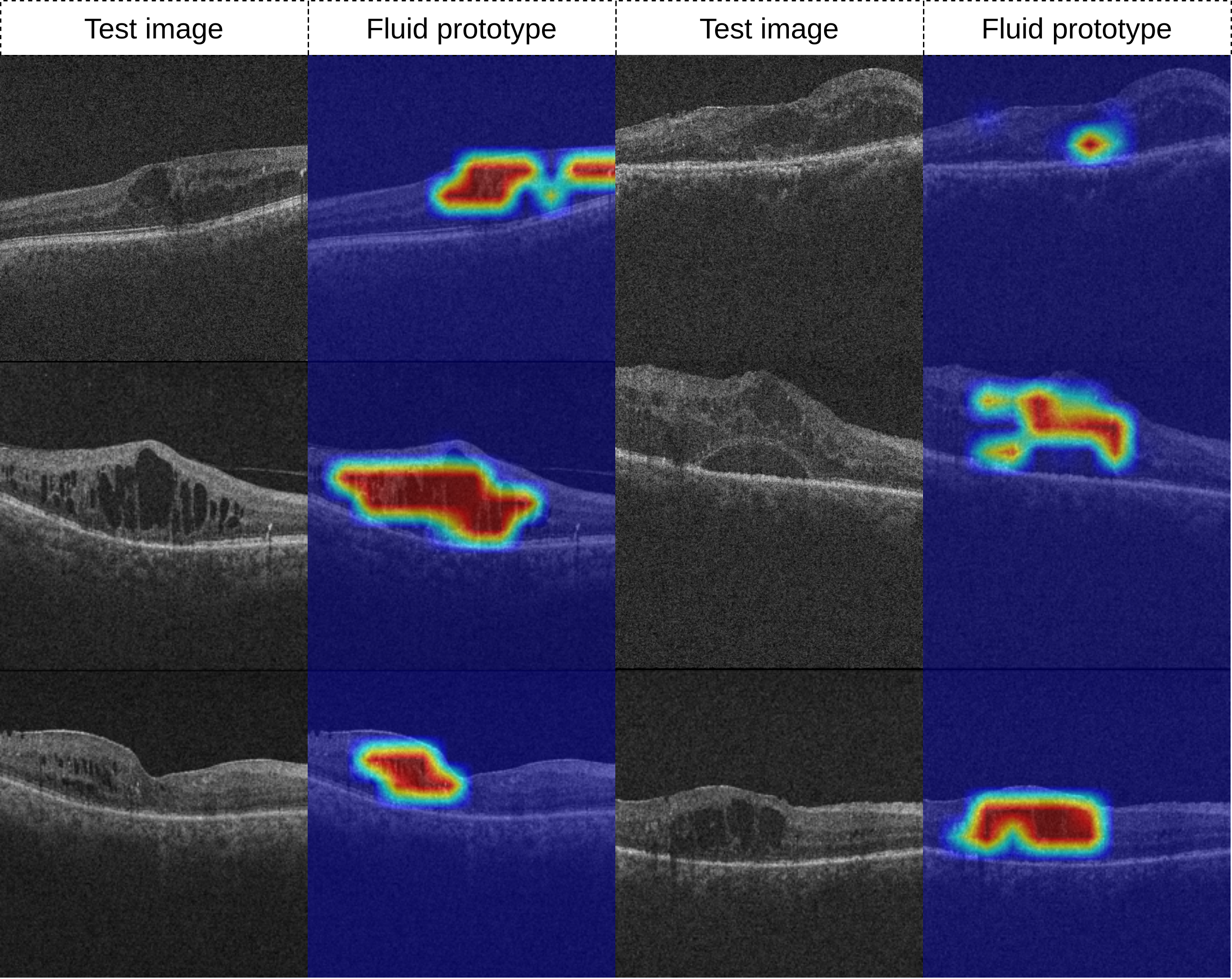}
    \caption{Visualization of the prototype corresponding to the fluid learned by PiPViT for class DME in OPTIMA5C.}
    \label{fig:pipvit_motivation}
\end{figure}

\begin{figure}[]
    \centering
    \includegraphics[width=\columnwidth]{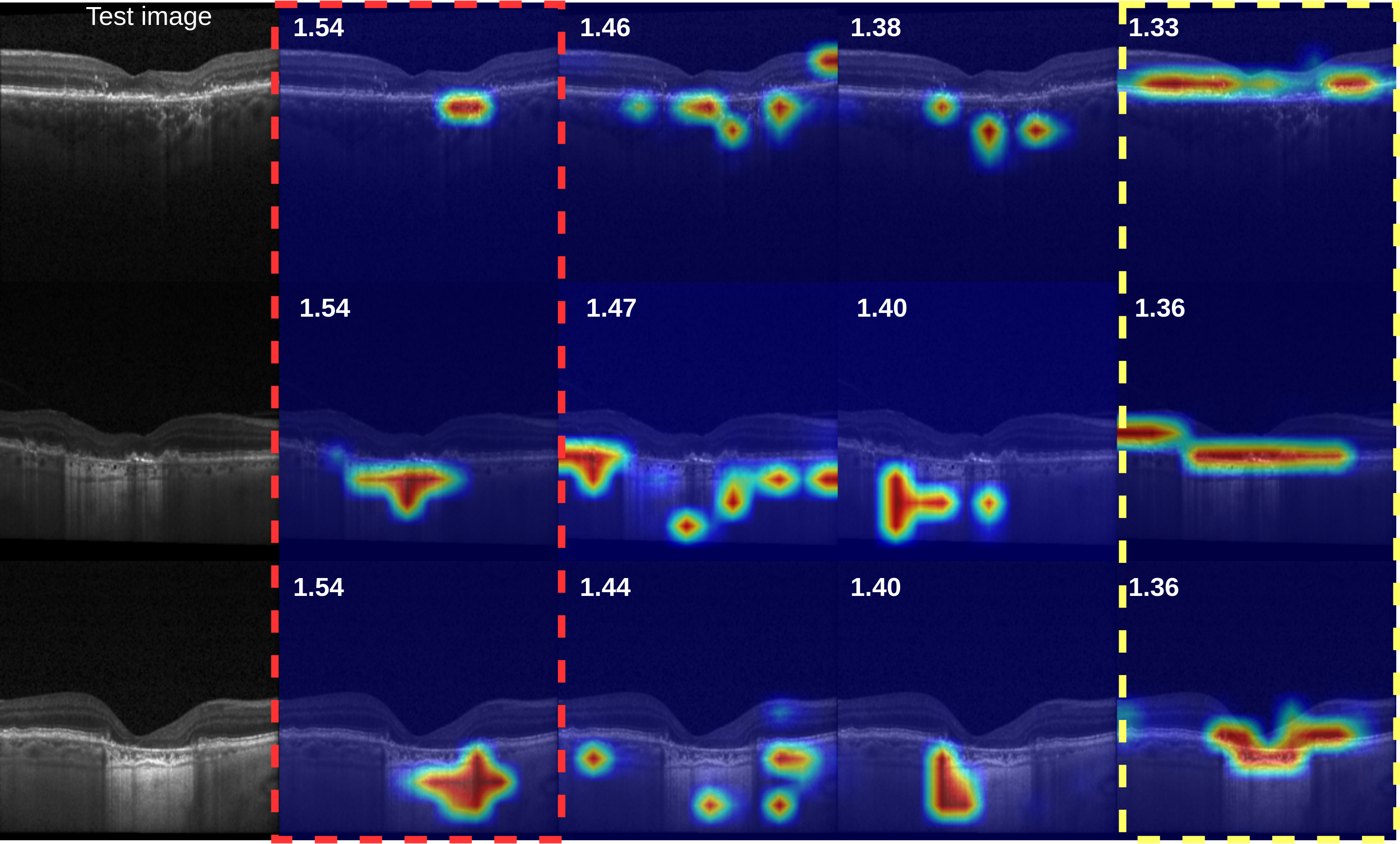}
    \caption{Visualization of the top-5 highly activated prototypes learned by PiPViT for class GA in OPTIMA5C.
    PiPViT effectively captures key GA biomarkers, including hyperreflective areas (column within the red rectangle) and retinal layer atrophy (column within the yellow rectangle).}
    \label{fig:ga_protoes}
\end{figure}

\begin{figure}[]
    \centering
    \includegraphics[width=\columnwidth]{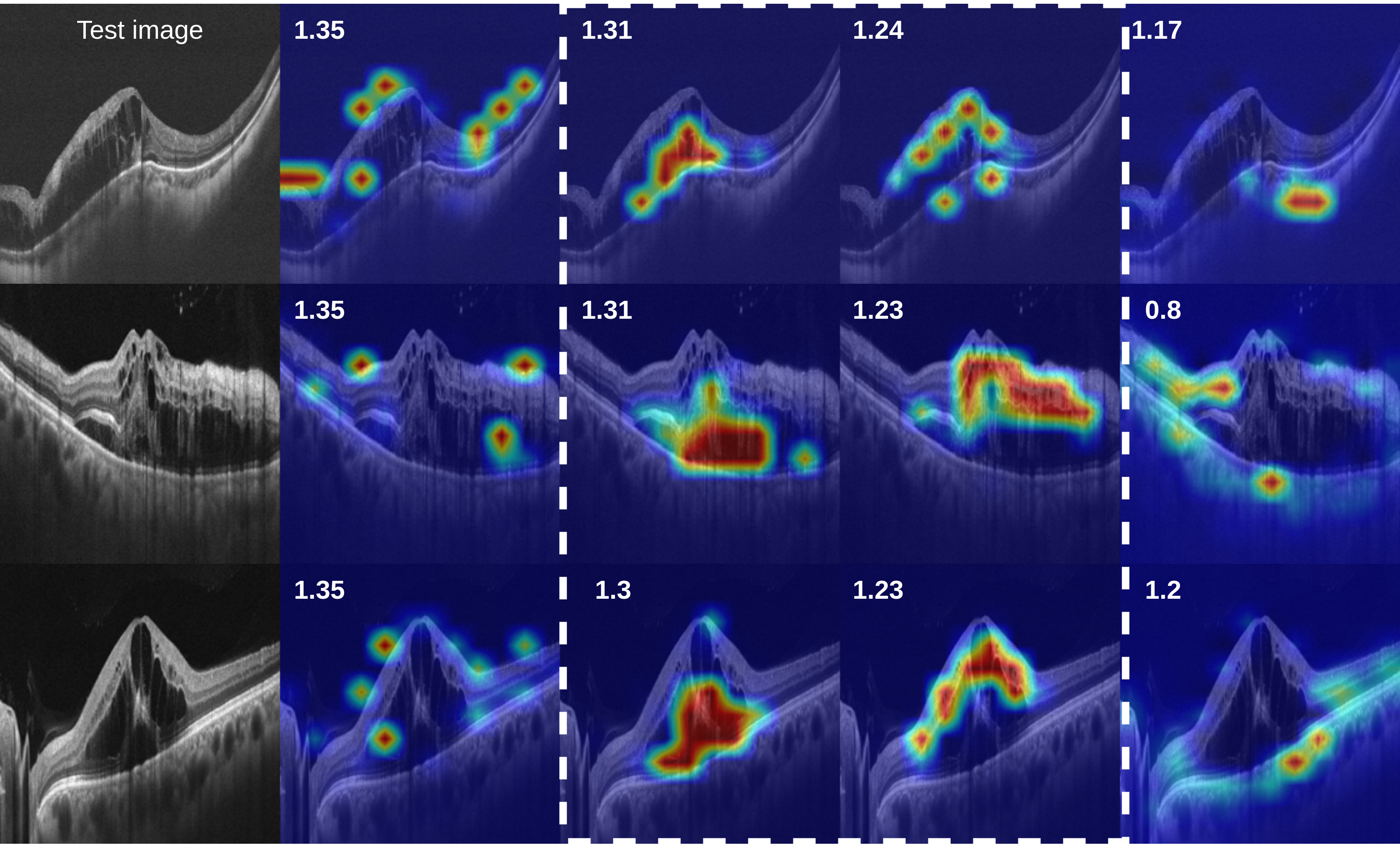}
    \caption{Visualization of the top-5 highly activated prototypes learned by PiPViT for class RVO in OCTDL.
    The second and third prototypes, highlighted within the white rectangle, align with macular edema, which typically manifests as cystic, localized swelling in the inner retina due to leakage from dilated veins.}
    \label{fig:rvo_protoes}
\end{figure}

Figures \ref{fig:dme_heatmaps} to \ref{fig:rvo_protoes} show explainability maps for various retinal diseases, comparing black-box, non-prototypical, and prototypical approaches.
Based on the qualitative results of post-hoc approaches, we draw the following conclusions:

\begin{itemize}
    \item Grad-CAM applied to ResNet-50 yielded coarse-grained maps that overlooked fine biomarker details and frequently failed to highlight critical regions (e.g., ERM in Fig.~\ref{fig:erm_heatmaps} (a)). However, it was effective for simpler tasks, such as distinguishing drusen from a normal retina by roughly localizing drusen and the surrounding area (Fig.~\ref{fig:drusen_heatmaps} (a)).
    \item Rollout showed inconsistent performance, with substantial noise and failure to highlight relevant image regions (Fig.~\ref{fig:dme_heatmaps} (a), Fig.~\ref{fig:erm_heatmaps} (a)). However, when applied to B-cos ViT, it highlighted meaningful regions for the corresponding disease, showing the potential of weight-input alignment.
    \item LRP on ViT produced less noisy maps than rollout but still missed crucial details (Fig.~\ref{fig:dme_heatmaps} (a)).
    \item The combination of the SWIN transformer and Score-CAM produced more explainable maps compared to other post-hoc methods. However, the generated maps remained coarse-grained (Fig.~\ref{fig:drusen_heatmaps} (a)) and typically highlighted only a single important biomarker at a time (Fig.~\ref{fig:erm_heatmaps} (a)).
\end{itemize}

From the prototypical models, we draw the following observations:

\begin{itemize}
    \item PiPViT and PiP-Net learned finer-grained, semantically meaningful prototypes, unlike traditional prototypical networks, which produced ambiguous, repetitive prototypes that often highlighted the same spatial region (see ProtoTree in Fig.~\ref{fig:dme_heatmaps} (b) and ProtoPFormer in Fig.~\ref{fig:erm_heatmaps} (b)).
    This issue likely arises from traditional prototype-learning approaches that constrain the network to summarize diverse concepts into a limited number of prototypes, thereby reducing semantic clarity.
    \item Although ProtoPFormer achieved high classification performance, its learned prototypes exhibited a semantic gap relative to the actual biomarkers (Fig.~\ref{fig:erm_heatmaps} (b)). Nevertheless, it occasionally learned more interpretable prototypes compared to traditional prototypical networks (Fig.~\ref{fig:dme_heatmaps} (b)). For drusen detection (Fig.~\ref{fig:drusen_heatmaps} (b)), however, it highlighted irrelevant regions, whereas ProtoPNet, despite its coarse focus, localized small drusen more accurately. This aligns with our expectations, as CNN-based models have a smaller field of view than ViT. Notably, PiPViT overcame this issue by multi-resolution pre-training, enhancing its ability to capture fine-grained biomarkers.
    \item While PiP-Net achieved high classification accuracy across datasets, its prototypes did not always align with human-interpretable biomarkers. For instance, on the OCTDL dataset, PiP-Net failed to learn a prototype for ERM (Fig.~\ref{fig:erm_heatmaps}~(b)).
    \item The comparison between PiPViT and PiP-Net prototypes in Fig.~\ref{fig:dme_heatmaps}~(b) highlights the advantage of global spatial attention in the ViT backbone, allowing PiPViT to more accurately capture the lesion extent than the CNN-based PiP-Net. 
    Notably,  as shown in Fig.~\ref{fig:pipvit_motivation}, the fluid prototype learned by PiPViT for the DME class in OPTIMA5C effectively approximates the extent of the fluid pocket—a key biomarker. In contrast, PiP-Net’s reliance on localized features resulted in overly fine-grained activations (Fig.~\ref{fig:PiP-Net_motivation}).
    \item PiPViT preserved feature diversity and identified key biomarkers. For example, the first three prototypes in Fig.~\ref{fig:dme_heatmaps}~(b) correspond to a swollen retina, retinal pigment epithelium changes, and intra-retinal fluid in DME. This alignment with human-understandable biomarkers is further confirmed by prototypes of GA in OPTIMA5C (Fig.~\ref{fig:ga_protoes}) and of RVO in OCTDL (Fig.~\ref{fig:rvo_protoes}), demonstrating PiViT's ability to capture clinically meaningful patterns.

\end{itemize}

\subsection{Quantitative Evaluation of Prototypes}\label{sec:drusen_detection}

Table \ref{tab:drusen-table} presents the results of the quantitative evaluation of drusen prototypes on the OCT5K dataset along with the classification AUC for the binary classification task (drusen versus Normal).

\begin{figure}[h]
    \centering
    \includegraphics[width=\columnwidth]{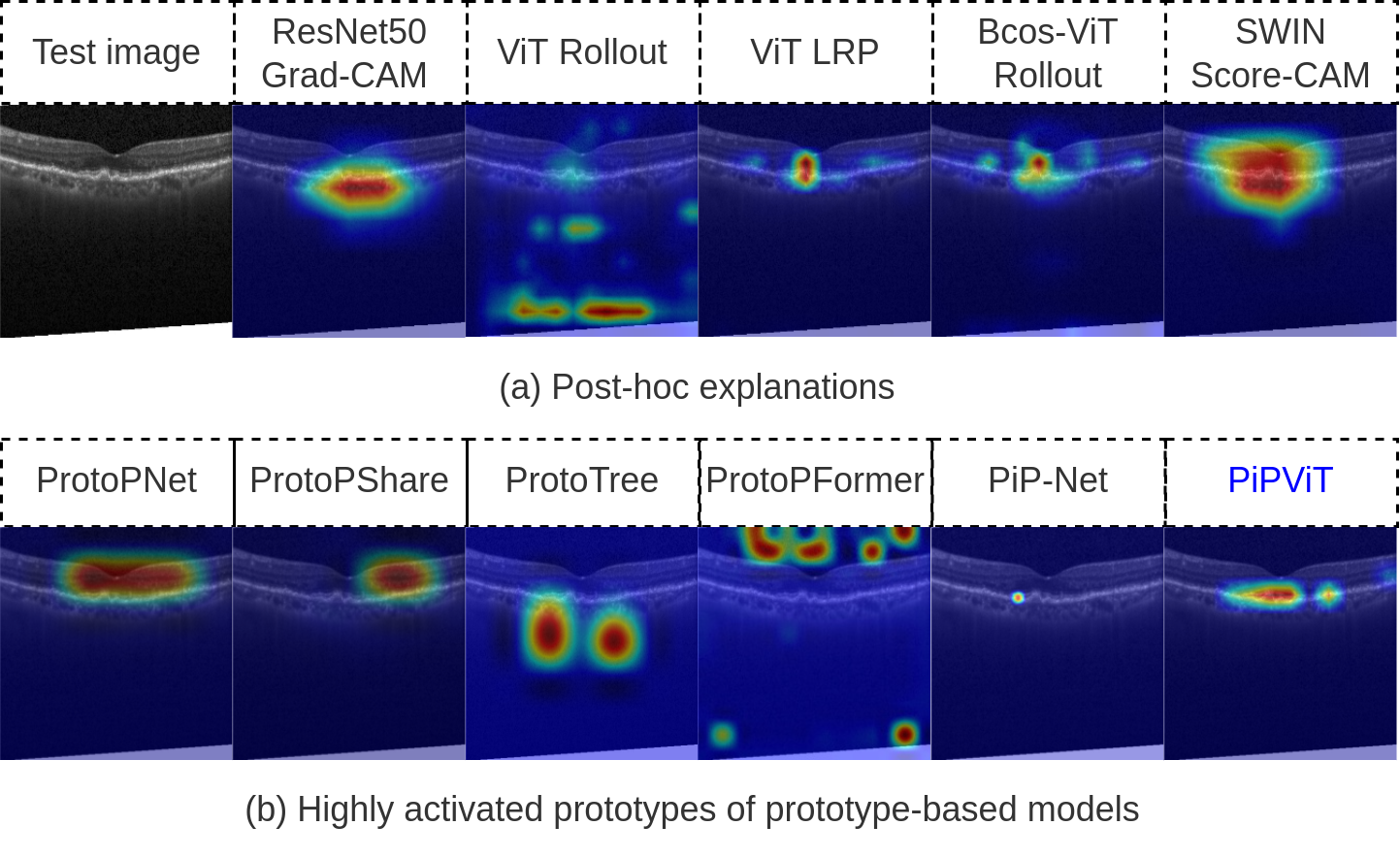}
    \caption{Visual explanation of different models' decisions for a drusen sample from OCT5K.
    (a) Heatmaps generated by post-hoc approaches applied to black-box and non-prototypical models.
    (b) Highly activated prototype of PiPViT vs. baseline prototypical methods.}
    \label{fig:drusen_heatmaps}
\end{figure}

Although traditional prototypical models (e.g., ProtoTree) achieved the highest classification performance, both qualitative (Fig.~\ref{fig:drusen_heatmaps}) and quantitative (Table \ref{tab:drusen-table}) evaluations showed that the learned prototypes did not semantically align with the drusen concept. In contrast, PiPViT produced finer-grained maps that accurately delineated drusen while excluding non-informative regions. PiPViT achieved an AP of $0.227$—significantly higher than ProtoTree's AP score of $0.1$.

Among traditional prototypical networks, ProtoPNet achieved a higher AP than PiPViT; however, the visualization of ProtoPNet's prototypes revealed coarse-grained activation maps that failed to precisely delineate the exact boundaries of drusen within the B-scan (see Fig. \ref{fig:drusen_heatmaps} (b)). Moreover, adding two highly activated drusen prototypes did not improve ProtoPNet’s performance in the drusen detection experiment, whereas it boosted PiPViT’s AP by $10.13$\%. This observation confirms that ProtoPNet’s prototypes were repetitive and focused on the same spatial region, while PiPViT learned higher-quality semantic prototypes that captured diverse drusen types.

In this experiment, despite its high classification performance (Table \ref{tab:overall}), PiP-Net struggled to capture the full extent of drusen lesions because its activation maps were overly fine-grained (Fig.~\ref{fig:drusen_heatmaps}(b)). This suggests that PiP-Net primarily learned the most representative lesion features while overlooking the complete lesion extent.

\begin{table}[]
\caption{Quantitative results of drusen detection on OCT5K dataset and binary classification performance (drusen vs. normal) on a subset of the Kermany dataset. We report the average precision (AP) when using the top-1 and top-2 most important prototypes that conceptualized drusen. Notably, PiP-Net learns only a single prototype for the drusen class.}
\label{tab:drusen-table}
\resizebox{\columnwidth}{!}{%
\begin{tabular}{@{}lcccccc@{}}
\toprule
\textbf{Metric \(\uparrow\)} & \textbf{\shortstack{Proto-\\PNet}} & \textbf{\shortstack{Proto-\\PShare}} & \textbf{ProtoTree} & \textbf{\shortstack{Protop-\\Former}} & \textbf{PiP-Net} & \textbf{\textcolor{blue}{PiPViT}} \\ \midrule
\textbf{AUC}                & 1              & 1     & 1     & 1     & \shortstack{0.994 \\ (0.988-0.998)} & \shortstack{0.99 \\ (0.98-0.998)} \\
\textbf{AP-top1}  & {0.272} & 0.158 & 0.100 & 0.001 & 0.023               & {0.227}       \\
\textbf{AP-top2} & {0.273} & 0.158 & 0.068 & 0.001 & -                   & {0.25}       \\ \bottomrule
\end{tabular}%
}
\end{table}
\subsubsection{Sensitivity Analysis of PiPViT's Pre-training on the Semantic Quality of Prototypes}

\begin{figure*}[t]
    \centering
    \includegraphics[width=0.95\textwidth]{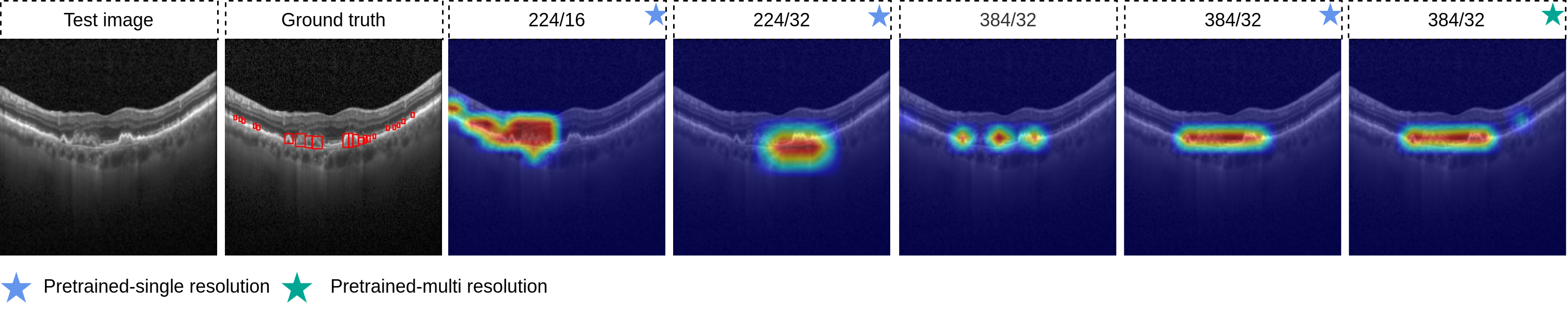}
    \caption{Visual comparison of the top-1 highly activated drusen prototype across different pre-training settings of PiPViT on OCT5K. We study the impact of image resolution, patch size, and incorporation of multi-resolution pre-training on the semantic quality of the learned drusen prototype.}
    \label{fig:ablation}
\end{figure*}

Table \ref{tab:resolution_ablation} and Figure \ref{fig:ablation} illustrate the impact of pre-training weights and multi-resolution pre-training strategy on the semantic quality of prototypes in PiPViT. 
The results indicate that increasing both patch size and input resolution enhanced drusen prototype's localization performance. Specifically, with a fixed input size ($224$), a larger patch size improved the AP for $10.79$\% and for a fixed patch size ($32$), a higher input resolution also led to an improvement of $7.18$\% in AP. 
On the other hand, multiscale pre-training led to an $8.61$\% improvement in AP compared to single-resolution pre-training. Additionally, it outperformed both single-resolution pre-training and training from random weight initialization.

\begin{table}
\caption{Sensitivity analysis of PiPViT for image resolution, patch size, and pre-trained weights on prototype semantic quality in OCT5K. Note that the underlined value indicates the final proposed setting.}
\label{tab:resolution_ablation}
\resizebox{\columnwidth}{!}{%
\begin{tabular}{@{}ccccc@{}}
\toprule
\textbf{Resolution} & \textbf{Patch size} & \textbf{\shortstack{Pre-trained \\ single-resolution}} & \textbf{\shortstack{Pre-trained \\ multi-resolution}} & \textbf{AP-top1 \(\uparrow\)} \\ \midrule
224 & 16 & \cmark  & \xmark & 0.176 \\
224 & 32 & \cmark  & \xmark & 0.195 \\ \midrule
384 & 32 & \xmark & \xmark & 0.166 \\
384 & 32 & \cmark  & \xmark & 0.209 \\
384 & 32 & \xmark & \cmark  & \ul{0.227} \\ \bottomrule
\end{tabular}%
}
\end{table}

\section{Conclusion and Future Work}
 
In this work, we propose PiPViT, a prototypical network for learning retinal biomarkers. Leveraging a ViT backbone, PiPViT learns semantically meaningful prototypes that can estimate lesion extent. Additionally, contrastive learning and multi-resolution pre-training enhance its ability to learn generalizable and robust features.
Experiments on five retinal OCT datasets show that PiPViT achieves competitive classification accuracy and delivers transparent, human-understandable decisions. For example, the quantitative evaluation of drusen prototypes learned by PiPViT showed that these prototypes are both discriminative and well-aligned with actual biomarkers, despite training only on image-level labels.

However, PiPViT faces some limitations. Although it effectively identifies key clinical features, its scoring mechanism may not always prioritize the most critical biomarkers, occasionally overemphasizing regions less clinically relevant for retinal diseases (e.g., the vitreous). Restricting the analysis to retinal layers could mitigate this issue, but would require extra segmentation and may not suit disease that affect the area outside the retina.

A promising direction for future improvement is to incorporate a teacher-student self-supervised feature distillation approach \cite{wei2022contrastive} to enhance contrastive learning. While contrastive learning primarily aligns patch-level features, the teacher-student framework can distill features from augmented views of the same image, yielding diverse representations that are more robust for downstream tasks. We believe that concurrently optimizing both patch-level and global-level representations can lead to more distinctive and meaningful prototype weights.
Overall, PiPViT represents a promising step toward inherently interpretable clinical models, complementing human expertise with transparent insights into retinal diseases.

\section{Statements of ethical approval}
Ethics approval for this post-hoc analysis was obtained from the Ethics Committee at the Medical University of Vienna (EK: 1246/2016).

\section{CRediT authorship contribution statement}
\textbf{Marzieh Oghbaie:} Writing – review \& editing, Writing – original draft, Visualization, Validation, Supervision, Software, Resources, Project administration, Methodology, Investigation, Formal analysis, Conceptualization. 

\textbf{Teresa Ara\'ujo:} Writing – review \& editing, Writing – original draft, Supervision, Methodology, Formal analysis, Conceptualization. 

\textbf{Hrvoje Bogunovi\'c:} Writing – review \& editing, Supervision, Resources, Project administration, Methodology, Funding acquisition, Conceptualization.

\section{Declaration of competing interest}
The authors declare that the research was conducted in the absence of any commercial or financial relationships that could be construed as a potential conflict of interest.

\section{Acknowledgements}
This work was supported in part by the Christian Doppler Research Association, Austrian Federal Ministry  of Economy, Energy and Tourism, the National Foundation for Research, Technology and Development, and Heidelberg Engineering.
\bibliographystyle{plainnat} 
\bibliography{main.bib}
\end{document}